\PassOptionsToPackage{hyphens}{url} 
\documentclass[11pt]{article}

\usepackage[utf8]{inputenc}
\usepackage[T1]{fontenc}
\usepackage{microtype} 
\usepackage{times} 
\usepackage{amsmath, amssymb, amsfonts}
\usepackage{graphicx}
\usepackage{lineno} 
\usepackage[margin=1in]{geometry} 
\usepackage{hyperref} 
\usepackage{caption}
\usepackage{authblk}
\usepackage{enumitem} 
\usepackage{booktabs} 
\usepackage{listings}
\usepackage{xurl} 

\hypersetup{
    colorlinks=true,
    linkcolor=blue,
    filecolor=magenta,
    urlcolor=cyan,
    citecolor=blue,
}

\lstdefinestyle{json}{
    basicstyle=\ttfamily\tiny, 
    numbers=left,
    numberstyle=\tiny\color{gray},
    stepnumber=1,
    numbersep=5pt,
    showstringspaces=false,
    breaklines=true,
    frame=lines,
    backgroundcolor=\color{white},
    stringstyle=\color{red!60!black},
    keywordstyle=\color{blue},
    commentstyle=\color{green!60!black},
    morestring=[b]",
    literate=
     *{0}{{{\color{black}0}}}1
      {1}{{{\color{black}1}}}1
      {2}{{{\color{black}2}}}1
      {3}{{{\color{black}3}}}1
      {4}{{{\color{black}4}}}1
      {5}{{{\color{black}5}}}1
      {6}{{{\color{black}6}}}1
      {7}{{{\color{black}7}}}1
      {8}{{{\color{black}8}}}1
      {9}{{{\color{black}9}}}1
      {\{}{{{\color{blue}\lbrace}}}1
      {\}}{{{\color{blue}\rbrace}}}1
      {[}{{{\color{blue}[}}}1
      {]}{{{\color{blue}]}}}1
      {:}{{{\color{black}:}}}1
      {,}{{{\color{black},}}}1,
}

\title{Modeling the Diachronic Evolution of Legal Norms: \\An LRMoo-Based, Component-Level, Event-Centric Approach to Legal Knowledge Graphs}

\author[1]{Hudson de Martim}
\affil[1]{Federal Senate of Brazil \\ \texttt{hudsonm@senado.leg.br}}
\date{} 

\begin{document}

\maketitle

\begin{abstract}
Representing the temporal evolution of legal norms is a critical challenge for automated processing. While foundational frameworks exist, they lack a formal pattern for granular, component-level versioning, hindering the deterministic point-in-time reconstruction of legal texts required by reliable AI applications. This paper proposes a structured, temporal modeling pattern grounded in the LRMoo ontology. Our approach models a norm's evolution as a diachronic chain of versioned F1 Works, distinguishing between language-agnostic Temporal Versions (TV) — each being a distinct Work — and their monolingual Language Versions (LV), modeled as F2 Expressions. The legislative amendment process is formalized through event-centric modeling, allowing changes to be traced precisely. Using the Brazilian Constitution as a case study, we demonstrate that our architecture enables the exact reconstruction of any part of a legal text as it existed on a specific date. This provides a verifiable semantic backbone for legal knowledge graphs, offering a deterministic foundation for trustworthy legal AI.
\end{abstract}


\section{Introduction}
The increasing demand for transparency, efficiency in legal research, and the proliferation of Legal Tech and Artificial Intelligence (AI) applications underscore the critical need for machine-readable representations of legal norms. In recent years, the development of robust Legal Knowledge Graphs (LKGs) has emerged as a key strategy to meet this need \cite{louis2024interpretable,moens2007automatic}. The challenge of formally modeling time and identifying legal resources through an event-centric ontology was a foundational topic in seminal early research \cite{lima2008ontology}, and it continues to be an active area of investigation \cite{kucuk2025computational,chen2023temporal,cai2024survey}.

While foundational conceptual frameworks like IFLA LRMoo \cite{lrmoo2024} and standards like Akoma Ntoso \cite{palmirani2011akoma} provide robust, general-purpose approaches, the specific challenge of tracking temporally discrete, component-level changes requires a dedicated modeling pattern. This challenge is particularly acute when considering current Generative AI. Brazil’s 1988 Federal Constitution, for instance, has been amended over one hundred times. Today's large language models (LLMs) operate on a probabilistic basis, making them inherently unsuitable for tasks requiring absolute precision. Their lack of deterministic mechanisms makes them unreliable for the high-stakes legal domain, where correctness is not a probability but a requirement \cite{padiu2024extent,lai2023large}. They frequently fail when asked to retrieve the exact text of the Constitution as it stood on a particular past date.

Against this backdrop, the present paper proposes a structured model grounded in the \textbf{IFLA LRMoo} ontology to deliver a semantically rich representation of legal norms and their diachronic evolution. We model this evolution as a chain of versioned \texttt{F1 Works}, distinguishing between a language-agnostic \textbf{Temporal Version (TV)}, each being a distinct F1 Work, and its concrete monolingual \textbf{Language Versions (LV)}, modeled as F2 Expressions. This paradigm is extended to the component level by establishing parallel hierarchies for abstract Components and their Component Versions. Furthermore, we formalize the amendment process itself through a scalable hierarchy of creation and end-of-existence events (anchored in \texttt{F27 Work Creation}), from which each version's validity is deterministically derived. This fine-grained, event-centric architecture forms the basis for building deterministic pipelines capable of reconstructing any part of a legal text as it existed on any given date.

This paper is structured as follows. Section \ref{fundamental_concepts} reviews foundational concepts from LRMoo and related standards. Section \ref{legal_norm_representation} introduces our core proposal for norms, detailing the TV/LV distinction and the event-centric mechanism. Section \ref{legal_norm_component_representation} details how this model is extended to represent internal hierarchical components. Section \ref{sec:conclusion_future_work} concludes and outlines future research directions.

\section{Background and Related Work}
\label{fundamental_concepts}
\subsection{LRMoo Overview}
The IFLA Library Reference Model (LRM), formalized as the LRMoo ontology \cite{lrmoo2024}, is an object-oriented model harmonized with the CIDOC Conceptual Reference Model (CIDOC CRM). It refines the Work, Expression, Manifestation, Item (WEMI) distinction:
\begin{itemize}
\item \textbf{F1 Work}, representing a distinct abstract intellectual creation (e.g., the normative intent of a law).
\item \textbf{F2 Expression}, representing a specific intellectual realization of a \emph{Work}. Differences in form imply different Expressions (e.g., from text to spoken word, a transcript of a recording). Similarly, differences in language imply different Expressions.
\item \textbf{F3 Manifestation}, representing the embodiment of an \emph{Expression} in a product (e.g., a PDF file published in the Official Gazette).
\item \textbf{F5 Item}, representing a single exemplar of a \emph{Manifestation} (e.g., a specific printed copy).
\end{itemize}

\subsection{Related Standards: Akoma Ntoso, LexML Brazil, and ELI}
\textbf{Akoma Ntoso (AKN)} \cite{palmirani2011akoma} is an XML vocabulary for legal documents aligned with the WEMI model. AKN does offer rich amendment constructs---\texttt{activeModifications} and \texttt{passiveModifications}, whose \texttt{textualMod} elements reference the affected provisions (by \texttt{eId}) and the amending source---so a textual change can be linked to the component it affects. These are, however, document-embedded XML annotations rather than first-class, reified entities: AKN does not model each component as an independently versioned work with an explicit derivative lineage, nor the amendment as a shared causal node, and point-in-time states are typically materialized as separate consolidated documents, leaving the per-component diachronic history implicit.

The \textbf{LexML Brazil} project \cite{lexml2008modelo} provides a URN standard (\texttt{norm@version\textasciitilde language!part}, implementing the URN Lex namespace) able to name a specific provision within a specific version of a norm---an addressing scheme we reuse to identify our entities. Its XML layer additionally records legislative events (e.g., amendments and partial revocations) as document metadata, and its FRBRoo-based ontology \cite{lima2008ontology} pioneered the time-aware modeling of the normative act's life cycle. A URN, however, \emph{identifies} a versioned provision as a locator within a consolidated document snapshot; it does not \emph{model} it as a first-class Work bearing explicit derivative lineage and conceptual membership, nor does it reify the legislative event as a queryable causal entity\footnote{In LexML the same amendment is recorded redundantly---actively in the amending norm's \texttt{EventosGerados} and passively in the amended norm's \texttt{CicloDeVida}. The two records are document-scoped: an \texttt{Evento}'s identifier is an \texttt{xsd:ID}, unique and resolvable only within its own file, whereas LexML reserves global identifiers (URN-LEX) for norms and provisions, not for events. Reusing the same id string across both documents is therefore a naming convention, not a resolvable co-reference, and a processor still correlates the two views heuristically (by matching source, target, type, and date). Reifying the event as a single first-class node gives it one global identity referenced from both sides, making the correlation deterministic.}. Our model supplies exactly this formal, relational layer, elevating each component and each event to a first-class entity.

The \textbf{European Legislation Identifier (ELI)} \cite{eli2014standard} also builds upon FRBR principles and supports both temporal versioning (through point-in-time URI components) and language version differentiation (via language codes in the URI). Its metadata model further captures relevant temporal dimensions, such as validity and applicability dates, on which we build in Section~\ref{legal_norm_representation}. However, ELI is fundamentally a URI identification and metadata scheme. While its accompanying formal ontology is based on FRBR principles, its semantics are document-centric, designed to describe relationships between laws (e.g., eli:amends). It does not provide the built-in constructs to model the legislative event as a first-class entity, which is essential for tracing the precise provenance of a legal change.

Grounding our model in \textbf{IFLA LRMoo} supplies the formal ontology these schemes lack: a recursive, component-level versioning pattern coupled with reified, event-centric provenance that traces each modification to its source legislative instrument.

While recent studies emphasize the need for LKGs to act as a verifiable backbone 
for computational law \cite{boer2020legal,gangemi2005constructive}, they often presuppose the existence of a stable, versioned representation of the legal text itself. Our model provides exactly this versioned textual ground truth, a necessary prerequisite for any subsequent semantic modeling or logical reasoning.

\section{LRMoo-Based Representation of Legal Norms}
\label{legal_norm_representation}
Building on LRMoo, our model maps each stage of a norm's existence:
\begin{itemize}
    \item \textbf{Concept (C):} A single, immutable \texttt{F1 Work} represents the abstract legal norm in its entirety, persisting through time. For example, "The Brazilian Federal Constitution of 1988" as an overarching concept is one such entity. It serves as the permanent anchor for all its historical versions.

    \item \textbf{Temporal Version (TV):} An \texttt{F1 Work} representing the norm’s complete and distinct normative identity at a specific point in time—a ``semantic snapshot.'' Each TV is a separate Work because legislative amendments alter the norm's legal force, creating a new normative entity. All TVs are linked to the overarching concept via \texttt{R10 is member of}---related works that are considered to share a common concept. Their temporal lineage is captured via \texttt{R2 is derivative of}, which can be precisely qualified using \texttt{R2.1 has type} to denote a ``Temporal Succession''.

    \item \textbf{Language Version (LV):} An \texttt{F2 Expression} representing the concrete monolingual textual realization of a Temporal Version. By being multiply instantiated as an \texttt{E33 Linguistic Object}, an LV is formally described by its language (\texttt{P72 has language}) and its complete textual content (\texttt{P190 has symbolic content}). An LV \texttt{R3i realises} its parent Temporal Version.
     
    \item \textbf{Manifestation (\texttt{F3 Manifestation}):} Embodies a Language Version on a specific carrier, such as an HTML page on a government website. The detailed modeling of manifestations is considered out of scope for this work.
\end{itemize}

\begin{figure}[htbp]
\centering
\includegraphics[width=0.7\textwidth]{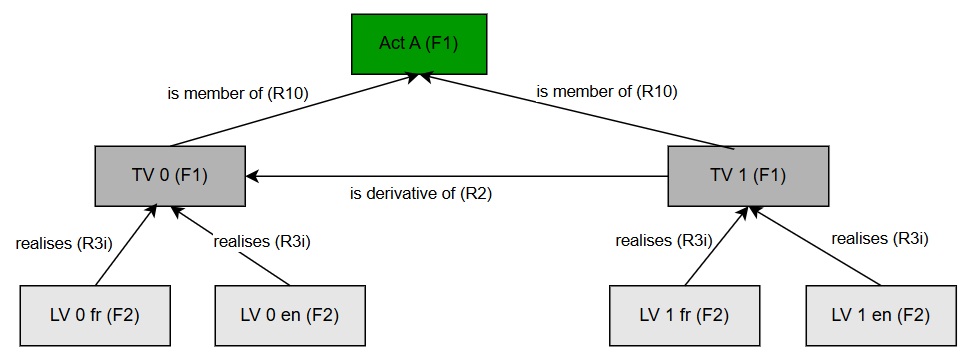} 
\caption{Relationship among the \texttt{Concept}, a chain of Temporal Versions (TVs) and their Language Versions (LVs).}
\label{fig:norm_versions_publications}
\end{figure}

This results in a clear structure, illustrated in Figure~\ref{fig:norm_versions_publications}.

The legislative enactment of an amending norm (e.g., through promulgation
or signing into law) has a creative effect resulting in at least two distinct 
\texttt{F1 Works}: the original version of the amending norm itself and the new version(s) of the amended norm(s). While the creation of the modifying instrument is a necessary outcome, the primary focus of this paper is the second, transformative effect: the creation of the new \textbf{Temporal Version (TV)} of an amended norm.

A legislative change is an inherently n-ary fact---it binds an author, a target norm, a resulting text, a date, and a legal effect---so collapsing it into a single binary relation (e.g., \texttt{amends}) would discard most of this structure \cite{noy2006defining}. We therefore reify the change as a first-class event, following event-centric knowledge graph approaches \cite{rospocher2016building,gottschalk2018eventkg}.

This transformative effect is modeled through a single event whose ontological typing depends on the nature of the legislative operation. For a textual amendment---the most common case---the event is multiply instantiated as both an \texttt{F27 Work Creation} and an \texttt{E64 End of Existence}, reflecting that a constitutive legal change simultaneously terminates one normative entity and inaugurates another.\footnote{We deliberately avoid modeling the change as an \texttt{E11 Modification}: the property \texttt{P31 has modified} ranges over physical things (\texttt{E24}/\texttt{E18}), whereas a Temporal Version is a conceptual \texttt{F1 Work} (\texttt{E28}). Modeling the transition at the level of the entity's \emph{existence} (\texttt{E63}/\texttt{E64}), rather than as a physical modification, keeps the pattern ontologically consistent. The combined \texttt{F27}\,$\cap$\,\texttt{E64} typing follows the CIDOC CRM precedent for state-transition events (e.g., a death jointly typed as \texttt{E64} and \texttt{E7}).} This event explicitly models the legal change by documenting that it:
\begin{itemize}
    \item \textbf{uses} (\texttt{P16 used specific object}) the specific provision of the amending instrument that prescribes the change, recording the authoritative source of the modification;
    \item \textbf{takes out of existence} (\texttt{P93 took out of existence}) the preceding Temporal Version (\texttt{TV\textsubscript{n-1}}) of the norm being changed, thereby terminating its validity period; and
    \item \textbf{creates} (\texttt{R16 created}) the new, resulting Temporal Version (\texttt{TV\textsubscript{n}}) that incorporates the normative changes.
\end{itemize}

\begin{figure}[htbp]
\centering
\includegraphics[width=0.6\textwidth]{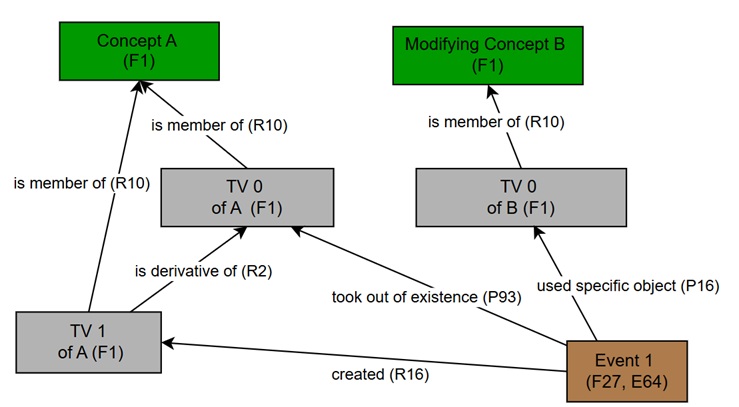} 
\caption{An amendment event, jointly typed as \texttt{F27 Work Creation} and \texttt{E64 End of Existence}: it \texttt{P16 used} the amending provision as source, \texttt{P93 took out of existence} the prior Temporal Version, and \texttt{R16 created} the new one.}
\label{fig:creation_event}
\end{figure}

The typing is operation-dependent: a pure insertion is typed only as \texttt{F27 Work Creation} (asserting \texttt{R16 created}); a pure repeal only as \texttt{E64 End of Existence} (asserting \texttt{P93 took out of existence}); and a wording amendment combines both, as above. This event-centric view allows us to precisely trace the provenance of each version of a legal norm, as shown in Figure~\ref{fig:creation_event}.

Essentially, a Temporal Version does not store its validity interval as a static attribute. Following an \emph{event-sourcing} approach, the interval $[t_{\text{start}}, t_{\text{end}})$ is \emph{derived} from the events bounding the version's life cycle: $t_{\text{start}}$ from the event that \texttt{R16 created} it, and $t_{\text{end}}$ from the event that \texttt{P93} took it out of existence---remaining open-ended while no such terminating event exists.

The event can be further described by its \textbf{Nature (\texttt{P2 has type})}, 
\textbf{Actors (\texttt{P14 carried out by})}, and 
\textbf{Time-span (\texttt{P4 has time-span})}.

The temporal dimension of legal knowledge is \textbf{bitemporal} in the strict technical sense \cite{snodgrass1999developing}: it requires distinguishing \emph{valid time}---when a norm produces legal effect---from \emph{transaction time}---when the norm was recorded in the official system. In our model, \texttt{P4 has time-span} captures transaction time (the publication/recording date), while valid time is carried by a \textbf{validity interval} attached to the event (an extension beyond those native to LRMoo). For legal corpora, valid time decomposes further: following the ELI ontology~\cite{eli2014standard} and the legal time model of Palmirani and Brighi~\cite{palmirani2006time}, we distinguish the date a norm \emph{enters the legal order} from its \textbf{applicability}---the temporal scope of the facts it may govern---because the two diverge under \emph{vacatio legis} (deferred onset), retroactive (\emph{ex tunc}) effect, and tax-anteriority or \emph{lex mitior} rules. A tax statute, for instance, may be published and in force in October yet only applicable to taxable events from the following January; conversely, a more lenient criminal law (\emph{lex mitior}) applies retroactively to facts predating its enactment. Modeling these as explicit properties on valid time renders such asymmetries computable rather than leaving them buried in unstructured text. The life-cycle interval $[t_{\text{start}}, t_{\text{end}})$ derived above is the valid-time projection; transaction time is recorded independently via \texttt{P4}.

The property \texttt{R2 is derivative of} is a semantic "shortcut" for this detailed event-based model, allowing for both simple version traversal and deep contextual analysis.

This architecture is designed to robustly handle both multilingual and 
monolingual scenarios. In jurisdictions like Canada or the European Union, 
a single language-agnostic \emph{Temporal Version} can be realized in 
multiple \emph{Language Versions}, one for each official language. In 
predominantly monolingual jurisdictions like Brazil, each TV will typically 
have one primary LV that realizes it. Even in this one-to-one case, 
the conceptual distinction remains essential: the TV defines \textbf{what the 
law says} (its semantic content and structure), while the LV defines 
\textbf{how it is expressed} (its specific linguistic form).

\subsection{Practical Application: Brazilian Constitution}
We trace the Constitution's lifecycle, using the LexML URN format (\texttt{norm@version \textasciitilde language} syntax)\footnote{LexML URN syntax follows the standard proposed by the Brazilian LexML project. Available at: \url{https://projeto.lexml.gov.br}} to identify the entities.

\paragraph{The Concept (F1)}
The abstract concept of the Constitution is a single \texttt{F1 Work}:
\begin{itemize}
\item \url{urn:lex:br:federal:constituicao:1988-10-05;1988}
\end{itemize}
\paragraph{Initial State (1988)}
Its promulgation created the initial \texttt{TV\textsubscript{0}} and its Portuguese \texttt{LV\textsubscript{0}}.
\begin{itemize}
\item \textbf{TV\textsubscript{0}}: \url{urn:lex:br:federal:constituicao:1988-10-05;1988@1988-10-05}
\item \textbf{LV\textsubscript{0}}: \url{urn:lex:br:federal:constituicao:1988-10-05;1988@1988-10-05~texto;pt}
\end{itemize}

\paragraph{Legislative Event (EC1-1992-03-31)}
Constitutional Amendment No. 1, promulgated on March 31, 1992, is modeled 
as a complex legislative event that creates \textbf{two} \texttt{F1 Works}:

\begin{itemize}
    \item \textbf{Work 1 -- The Amendment Instrument}: The normative act 
    EC No. 1/1992 itself, which legally prescribes specific modifications 
    to the constitutional text. This work has its own \texttt{F2 Expression} 
    containing the amendment's official text.
    
\item \textbf{Work 2 -- The Amended Constitution}: The new 
    \texttt{TV\textsubscript{1}} (1992-03-31), which represents the 
    constitutional framework incorporating the changes. This work is created 
    via an event jointly typed as \texttt{F27 Work Creation} and 
    \texttt{E64 End of Existence} that \textbf{uses} (\texttt{P16}) the relevant 
    provision of EC No. 1/1992 as its authoritative source, \textbf{takes out of 
    existence} (\texttt{P93}) the prior \texttt{TV\textsubscript{0}}, and 
    \textbf{creates} (\texttt{R16}) the new \texttt{TV\textsubscript{1}}.
\end{itemize}

\paragraph{New State (Post-Amendment)}
The event's result was a new Temporal Version, \texttt{TV\textsubscript{1}}, with its Language Version \texttt{LV\textsubscript{1}}.
\begin{itemize}
\item \textbf{TV\textsubscript{1}}: \url{urn:lex:br:federal:constituicao:1988-10-05;1988@1992-03-31}
\item \textbf{LV\textsubscript{1}}: \url{urn:lex:br:federal:constituicao:1988-10-05;1988@1992-03-31~texto;pt}
\end{itemize}
This establishes an auditable version chain: \texttt{TV\textsubscript{1} R2 is derivative of TV\textsubscript{0}}.

\section{Legal Norm’s Components Representation}
\label{legal_norm_component_representation}
Legal norms have a hierarchy of components (articles, paragraphs, etc.). Our model applies the same LRMoo pattern recursively to each component.

\begin{itemize}
    \item \textbf{Component Concept (CC):} An \texttt{F1 Work} 
    representing the abstract identity of a structural component of a legal norm 
    (e.g., "Article 5"). The hierarchy of CCs is structured using 
    \texttt{R67 has part}.
    
    \item \textbf{Component Temporal Version (CTV):} An \texttt{F1 Work} 
    representing the semantic content of a component at a specific point in time. 
    A CTV \texttt{R10 is member of} its corresponding CC. Amendments create new 
    CTVs, linked via \texttt{R2 is derivative of}.
    
    \item \textbf{Component Language Version (CLV):} An \texttt{F2 Expression} 
    representing the monolingual realization of a CTV. A CLV \texttt{R3i realises} 
    its corresponding CTV. Translations are linked via \texttt{R76 is derivative of}.
\end{itemize}

\begin{figure}[htbp]
\centering
\includegraphics[width=0.8\textwidth]{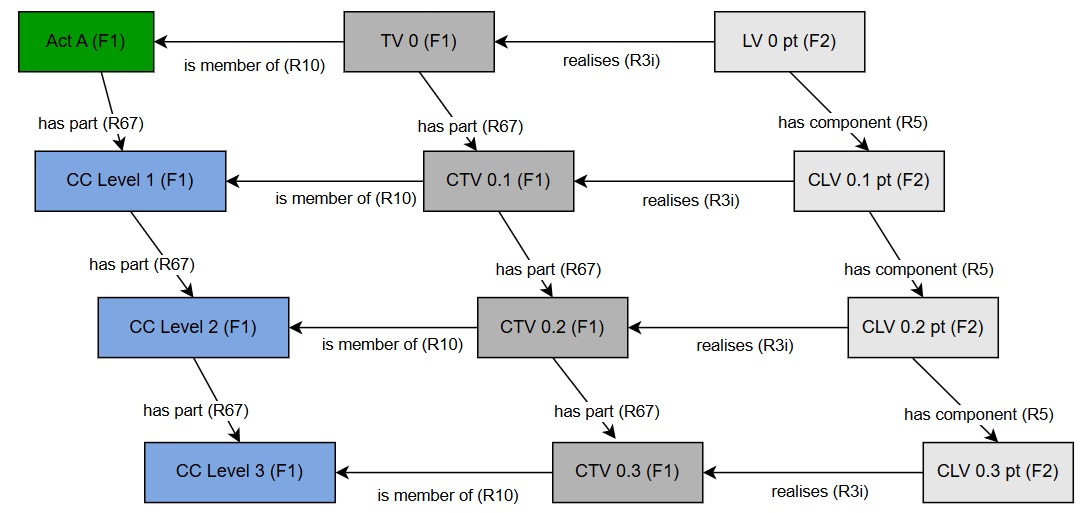} 
\caption{Parallel hierarchies of Works (structured by \texttt{R67 has part}) and Expressions (structured by \texttt{R5 has component}).}
\label{fig:component_hierarchy}
\end{figure}

The formal distinction between properties at each level, as showed in Figure \ref{fig:component_hierarchy}---\texttt{R10} 
(conceptual membership), \texttt{R67} (work composition), \texttt{R3} 
(work realization), and \texttt{R5} (textual incorporation)---is key to 
maintaining the model's semantic integrity and enabling precise querying 
across abstraction levels.

\subsection{Temporal Evolution and Event Granularity of Components}
Amendments are modeled with a scalable event hierarchy that leverages CIDOC CRM's mechanism for process decomposition. A single \textbf{macro-event} (e.g., "Enactment of an Amending Norm") captures the overall change and is formally composed of multiple \textbf{micro-events} using the \texttt{P9 consists of} property\footnote{Although the property \texttt{P9 consists of} is formally defined in CIDOC CRM for \texttt{E4 Period}, its extension, LRMOO, explicitly recommends its use for hierarchical relationships among activities. As \texttt{F27 Work Creation} is a subclass of \texttt{E7 Activity}, it inherits this applicability. The LRMOO documentation considers this use “more specific (and appropriate)” for decomposing an \texttt{F27} into component events (e.g., \texttt{F31 Performance} sub-activities), thus providing the rationale for its analogical application to decompose a macro-level creation event into its micro-level constituent parts.}. Each micro-event happens concurrently, representing a single, granular modification to a component's version.

Since the direct outcome of this granular action is a new semantic version of the component, each micro-event is typed dynamically according to its dogmatic operation, mirroring the pattern introduced in Section~\ref{legal_norm_representation}: a pure insertion is an \texttt{F27 Work Creation}, a pure repeal an \texttt{E64 End of Existence}, and a wording amendment is jointly typed as both. This creates an explicit, auditable link between the cause and effect of the change. Taking a wording amendment as the general case, the micro-event documents that it:

\begin{itemize}
    \item \textbf{uses} (\texttt{P16 used specific object}) the original Component Temporal Version (\texttt{CTV\textsubscript{0}})  of the specific provision of the amending instrument that prescribes the change, recording its authoritative source;
    \item \textbf{takes out of existence} (\texttt{P93 took out of existence}) the preceding Component Temporal Version (\texttt{CTV\textsubscript{n-1}}) of the component being changed, thereby terminating its validity period; and
    \item \textbf{creates} (\texttt{R16 created}) the new, resulting Component Temporal Version (\texttt{CTV\textsubscript{n}}) that incorporates the normative changes.
\end{itemize}
This hierarchical event structure is essential for establishing a complete auditable trail and data provenance. Its power is twofold. On one hand, the detailed record of the micro-event—as part of an overarching macro-legislative act—answers precisely \textit{how}, \textit{when}, and \textit{why} a component changed, tracing a specific instruction in an amending act to its precise effect on the amended norm. On the other hand, the outcome of this event creates a new Component Temporal Version (CTV) linked to its predecessor via \texttt{R2 is derivative of}. This property efficiently answers \textit{what} changed by forming a simple, traversable version history, while unaffected components simply retain their existing CTVs.  

While this hierarchical event structure provides a precise target for establishing a complete auditable trail, its instantiation at scale represents a significant practical challenge. The parsing of unstructured legislative texts to reliably extract the inputs and outputs for each micro-event is a non-trivial task, underscoring the importance of developing dedicated processing pipelines as part of future work.

\subsubsection{Practical Example: The Lifecycle of a Component}
We trace the caput of Article 6 of the Brazilian Constitution, which enumerates social rights.
\paragraph{The Component Concept (CC)}
The abstract concept of "the caput of Article 6" is a single \texttt{F1 Work}.
\begin{itemize}
\item \textbf{CC}: \url{urn:lex:br:federal:constituicao:1988-10-05;1988!art6_cpt}
\end{itemize}

\paragraph{The Initial State (1988)}
Its original 1988 text is captured in its first \texttt{CTV\textsubscript{0}} and Portuguese \texttt{CLV\textsubscript{0}}.
\begin{itemize}
\item \textbf{CTV\textsubscript{0}}: \url{urn:lex:br:federal:constituicao:1988-10-05;1988@1988-10-05!art6_cpt}
\item \textbf{CLV\textsubscript{0}}: \url{urn:lex:br:federal:constituicao:1988-10-05;1988@1988-10-05~texto;pt!art6_cpt}
\end{itemize}

\paragraph{The Component-Level Event (EC26-2000-02-14!art1\_cpt\_...)}
On Feb 14, 2000, the caput of Article 1 of Constitutional Amendment No. 26 added the word “housing” (“moradia”) to the caput of Article 6 of the Constitution. This specific instruction (action) is a micro-event. Being a wording amendment, it is jointly typed as an \texttt{F27 Work Creation} and an \texttt{E64 End of Existence}: it \textbf{uses} (\texttt{P16}) the \texttt{CTV\textsubscript{0}} of the amending provision (\url{urn:lex:br:federal:emenda.constitucional:2000-02-14;26@2000-02-14!art1_cpt_alt1_art6}) as its authoritative source, \textbf{takes out of existence} (\texttt{P93}) the preceding \texttt{CTV\textsubscript{0}} of the Art.~6 caput, and \textbf{creates} (\texttt{R16}) the new version, \texttt{CTV\textsubscript{1}}.

\paragraph{The New State (Post-Amendment)}
\begin{itemize}
\item \textbf{CTV\textsubscript{1}}: \url{urn:lex:br:federal:constituicao:1988-10-05;1988@2000-02-14!art6_cpt}
\item \textbf{CLV\textsubscript{1}}: \url{urn:lex:br:federal:constituicao:1988-10-05;1988@2000-02-14~texto;pt!art6_cpt}
\end{itemize}
The new Portuguese \texttt{CLV\textsubscript{1}} realises \texttt{CTV\textsubscript{1}}, containing the updated text. The lineage \texttt{CTV\textsubscript{1} R2 is derivative of CTV\textsubscript{0}} preserves the full, auditable history of this single component.

\section{Conclusion and Future Work}
\label{sec:conclusion_future_work}
This paper has proposed a structured, temporal model for representing legal norms and their hierarchical components, grounded in \textbf{IFLA LRMoo}. By representing temporal states as a chain of versioned \texttt{F1 Works}, our proposal offers a robust, standards-compliant solution for deterministic point-in-time reconstruction of legal texts. The key contributions are:
\begin{itemize}
    \item \textbf{A Two-Tier Expression Model:} A clear distinction between a language-agnostic \emph{Temporal Version (TV)} and its monolingual realizations, the \emph{Language Versions (LV)}, providing a precise mechanism for managing multilingual content.
    \item \textbf{Granular Component Versioning:} A recursive application of this model to internal components using parallel hierarchies for \emph{Component Concepts} and \emph{Component Versions}.
    \item \textbf{Event-Centric Change Modeling:} A scalable event hierarchy that traces an amendment from its precise provision in an amending norm to its exact effect on an amended norm, with each version's validity and applicability derived from these events.
    \item \textbf{Deterministic Reconstruction:} A formal structure that provides the unambiguous, verifiable ground truth essential for high-stakes legal applications, in direct contrast to probabilistic generative AI.
\end{itemize}

Future work will focus on three main areas:
\begin{itemize}
    \item \textbf{Full Ontological Implementation:} Develop and publish a comprehensive OWL ontology that formally specifies all entities and relationships, ensuring alignment with CIDOC CRM and LRMoo.
    \item \textbf{Knowledge Graph Population and Tooling:} Implement the model as a large-scale LKG. This involves developing automated pipelines to parse historical legislation and instantiate version and event entities, and creating tools for deterministic point-in-time reconstruction.
    \item \textbf{Advanced Reasoning and Benchmarks:} Build upon the LKG to develop advanced reasoning. A critical step will be creating benchmarks to rigorously test the model's ability to perform flawless, deterministic reconstruction.
\end{itemize}

In an era dominated by probabilistic models, a robust descriptive model such as this provides the precise, versioned textual facts—the \textbf{verifiable ground truth}—that are essential for building reliable Legal Knowledge Graphs. We frame this model as a foundational prerequisite for effective computational law, providing the deterministic and auditable factual layer required for trustworthy legal AI systems.

\bibliographystyle{ios1} 
\bibliography{referencias}

\end{document}